\documentclass[runningheads]{llncs}
\usepackage[T1]{fontenc}
\usepackage{tabularx}
\usepackage{booktabs}
\usepackage{array}
\usepackage{graphicx}
\begin{document}
\title{A Decade of Human–Robot Interaction Through Immersive Lenses: Reviewing Extended Reality as a Research Instrument in Social Robotics}

\titlerunning{A Decade of Human-Robot Interaction Through Immersive Lenses}
%
\author{André Helgert\orcidID{0000-0001-6008-4793} \and
Carolin Straßmann\orcidID{0000-0002-9473-2944} \and
Sabrina C. Eimler\orcidID{0000-0001-8944-2814}}
\authorrunning{A. Helgert et al.}
%
\institute{Institute of Computer Science and Institute of Positive Computing, University of Applied Sciences Ruhr West,
Lützowstrasse 5, 46236 Bottrop, Germany \\
\email{andre.helgert@hs-ruhrwest.de}}
\maketitle              
\begin{abstract}
Over the past decade, Extended Reality (XR), including Virtual, Augmented, and Mixed Reality, gained attention as a research instrument in human-robot interaction studies, but remains underexplored in empirical investigations of social robotics. To map the field, we systematically reviewed empirical studies from 2015 to 2025. Of 6,527 peer-reviewed articles, only 33 met strict inclusion criteria. We examined (1) how XR and virtual social robots are used, focusing on the software and hardware employed and the application contexts in which they are deployed, (2) data collection and analysis methods, (3) demographics of the researchers and participants, and (4) the challenges and future directions. Our findings show that social XR-HRI research is still driven by laboratory simulations, while crucial specifications – such as the hardware, software, and robots used – are often not reported. Robots typically act as passive and hardly interactive visual stimulus, while the rich biosignal (e.g., eye-tracking) and logging (e.g. motion capturing) functions of modern head-mounted displays remain largely untapped. While there are gaps in demographic reporting, the research teams and samples are predominantly tech-centric, Western, young, and male. Key limitations include hardware delays, small homogeneous samples, and short study cycles. We propose a four-phase roadmap to establish social XR-HRI as a reliable research medium, which includes (1) strengthen application contexts,  (2) more robust and testable technological iterations,  (3) embedding diversity in samples and research teams,  and (4) the need for reporting standards, e.g., in form of a suitable taxonomy. Advancing in these directions is essential for XR to mature from a lab prototype into an ecologically valid research instrument for social robotics.

\keywords{human-robot interaction\and social robotics\and virtual reality\and mixed reality\and augmented reality\and extended reality\and systematic literature review\and research methods.}
\end{abstract}
\newpage
\section{Introduction}
Immersive technologies, including Virtual Reality (VR), Augmented Reality (AR), and Mixed Reality (MR) – collectively referred to as Extended Reality (XR) – have been established over the past decade as powerful research tools~\cite{radianti2020systematic}. In the field of human–robot interaction (HRI), XR has been applied across a wide spectrum of scenarios ranging from controlled studies examining social interactions between humans and robots~\cite{Strassmann2024DontJudgeBook} to robot teleoperation~\cite{Meng.2023} and simulation settings where conducting experiments in the real world would be unsafe or impractical~\cite{Badia.2022,Arntz2022VIPER}. Moreover, the flexibility and adaptability of XR technologies are increasingly leveraged to design and evaluate custom robot prototypes~\cite{ge2023codesign}. Those virtual clones allow researchers to replicate a robot’s physical appearance and functional capabilities, while also enabling experimenting with prototypical designs or creative concepts that cannot yet be realized in physical robots due to technical or material limitations. The ability to rapidly modify robot design, behavior, and interaction modalities within virtual spaces shortens development cycles and enables studies that would be infeasible under real-world conditions~\cite{Coronado.2023}. XR provides researchers with a highly dynamic simulation platform for exploring and studying robotic systems. Beyond its technical advantages, XR provides a means of examining the social dimensions of HRI in an environment that is both experimentally controlled and ecologically valid~\cite{Parsons2015}. Prior work has shown that human behavior in virtual environments often mirrors behavior in physical contexts~\cite{Gromer.2018,Renison.2012}, which increases confidence in the validity of findings from XR-based experiments~\cite{Kisker.2021}. This makes XR particularly valuable in social robotics, where understanding human responses to robot gestures, speech, and other nonverbal cues is essential~\cite{Lei.2023}. Such interactions are not perceived solely on a functional level; rather, they activate human social schemas, prompting people to respond as if robots were social agents~\cite{rosenthal2013emotional,riek2008anthropomorphism}. Prior studies demonstrate that even subtle variations of factors such as a robot’s anthropomorphic features, emotional expressiveness, or communication style can strongly shape user perceptions and trust \cite{KLUBER2025100131}. Therefore, a need for experimental flexibility and controllability regarding both the robot and its interaction context is required. Within the domain of social robotics, XR uniquely addresses these needs~\cite{helgert2023virtualrealitytoolstudying}, enabling researchers to conduct studies without access to physical robots, design entirely novel robotic embodiments, and test experimental communication channels – such as alternative speech patterns, gestures, or facial expressions. Moreover, contextual variables such as the study environment can be systematically manipulated to investigate their influence on human perception and behavior. Despite these opportunities and the growing number of XR-HRI studies, research that uses XR as a research tool to investigate social HRI remains fragmented across different application contexts. The variety of (1) the conceptional roles of XR (e.g., simulation, prototyping, teleoperation, behavioral testing) and the interaction contexts in which social phenomena are investigated, (2) the methods and study design used to measure social cues, (3) the technical implementation and fidelity of virtual robots and environments, and (4) the composition of participant samples and research teams affects the comparability and generalizability of XR-HRI work. This heterogeneity makes it difficult to synthesize results, assess external validity, and derive a coherent picture of how XR is currently used to study social robots. To address this gap, this work presents a Systematic Literature Review (SLR) consolidating current knowledge, identifying recurring patterns, and outlining promising directions for future work on XR as a research tool in social robotics. From this objective, the following research questions are derived:

\medskip

\textbf{RQ1:} What roles and contexts do XR and virtual robots hold as a research tool for investigating social robots?

\medskip

\textbf{RQ2:} What data collection and analysis methods, as well as research designs, are used in studies that employ XR as a research tool for investigating social robots?

\medskip

\textbf{RQ3:} What challenges and future research directions can be identified in the field of XR as a research tool for social robots?

\medskip

\textbf{RQ4:} Which characteristics define the researchers and participants in studies that use XR technologies as a research tool for investigating social robots?

\section{Background of XR in Research}
XR has been established as a valid research instrument across various disciplines. For example, the medium is used in medicine as a training and simulation tool~\cite{Hanke2024}, in educational settings to display content more immersively~\cite{Disch}, and as an awareness-raising tool for a wide range of social issues and inequalities~\cite{catc,heilmann}. Although experiments on the fundamental idea of VR as we know it today were already conducted in the 1990s and early 2000s~\cite{NAP4761,biocca1992virtual}, XR only began to emerge as a factor in research from around 2015, following the release of advanced VR headsets from companies such as Facebook, HTC, and Sony. Since then, XR has evolved in terms of hardware, new application areas have been identified, and the advantages of XR as a research instrument have been recognized. These advantages include, for example, that virtual environments can present human simulations in a highly immersive way, creating a strong sense of presence. The concepts of presence and immersion in XR, identified by Rebelo et al.~\cite{Rebelo2012}, have since been considered key principles, not only to be maintained in XR experiences but also as important factors in understanding the physical and psychological states of participants. Furthermore, presence plays a key role in the fact that VR, in particular, is widely acknowledged to offer high ecological validity. This means that researchers can have a highly controllable experimental environment~\cite{Diemer2015}, while at the same time conducting studies in simulated field settings. It is also well established that people in VR display emotional responses similar to those in the real world~\cite{Gromer.2018} and exhibit comparable behavior to real-world settings~\cite{Renison.2012}. All these advantages result in the possibility of conducting cost-effective, location-independent, and valid research across various fields.


\newpage

\section{Materials and Methods}
This section reports the inclusion and exclusion criteria as well as the process of collecting relevant studies for this literature review following the PRISMA guidelines proposed by~\cite{moher2009prisma,page2021prisma2020}, as well as the coding scheme, which was used to systematically evaluate the relevant publications.

\subsection{Inclusion and Exclusion Criteria}

Studies were included if they addressed social HRI involving VR/AR/MR and explicitly employed these technologies as research instruments (e.g., simulation, experiment, or evaluation). Social HRI is defined as human–robot interaction that involves social communication or relationship-building processes (e.g., verbal/nonverbal cues, cooperation, trust, engagement), rather than purely functional task execution. Only empirical studies (e.g., experiments, user studies) published since January 2015 were considered because since then, mass-market VR-devices have been introduced. In addition, the articles had to be written in English and the full text of the article had to be accessible. Studies were excluded if they deviated from the topic of social robotics (e.g., industrial robots or no immersive medium) or if XR was only used for visualization without an experimental context. Furthermore, purely theoretical or technical articles without user evaluation were excluded, and care was taken to exclude outdated or non-generally available XR systems (experimental devices prior to 2015, e.g., Oculus Rift DK1).

\subsection{Identification of Articles}

\begin{figure*}[h]
    \centering
    \includegraphics[width=0.8\linewidth]{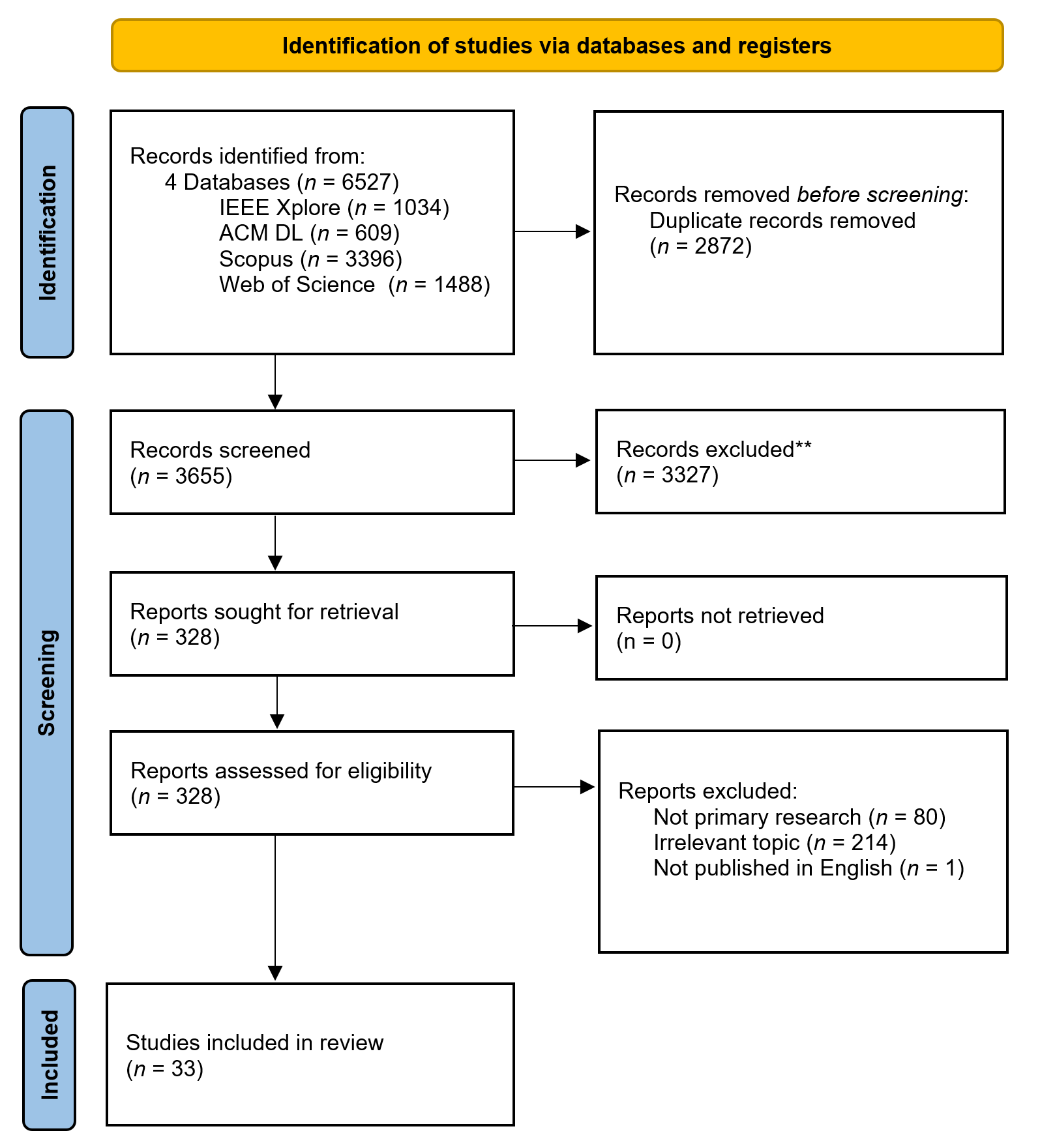}
    \caption{PRISMA flow chart of the search and selection process.}
    \label{fig:tut1}
\end{figure*}

The literature review was conducted within the databases IEEE Xplore, ACM Digital Library, Scopus, and Web of Science in April 2025, including articles published up to that date. The same search string was used in all databases, but adapted to the respective syntax: (("Human-Robot Interaction" OR "Robot" OR "HRI") AND ("Virtual Reality" OR "VR" OR "Mixed Reality" OR "MR" OR "Augmented Reality" OR "Extended Reality" OR "AR" OR "XR") AND ("Study" OR "Experiment" OR "Empirical" OR "Evaluation")). We ensured that the three core points – social HRI, XR, and empirical studies – were covered. Using this search string, 6,527 peer-reviewed articles were found. Of these, 2,872 duplicate records were removed. The screening comprised two stages: In the first stage, titles and abstracts were checked according to the criteria described above, whereby 3,327 articles were not off-topic; in the second stage, the full texts of the remaining contributions were reviewed. Out of these 328 articles, 33 final articles remained, which entered the analysis process. The excluded articles were not primary research (e.g. technical papers), irrelevant topics (e.g. industrial robots) or not published in English. The PRISMA flow diagram, following the implications of~\cite{page2021prisma2020}, is shown in Figure~\ref{fig:tut1}.

\subsection{Coding Scheme}

The unit of analysis was the individual publication. Coding was performed using a structured codebook, which was created deductively and contained operationalized definitions for all categories. For \textbf{RQ1}, codes were developed to classify the XR technology used, its overarching role, and the type of robot. XR is used in many settings, but it is rarely classified systematically. Cárdenas-Robledo et al. proposed a taxonomy in the context of Industry 4.0, noting that XR roles often overlap and are difficult to distinguish clearly~\cite{ADRIANACARDENASROBLEDO2022101863}. They point out that multidimensional perspectives are needed to adequately describe XR applications. In the absence of alternative taxonomies, we adapted the categorization of Cárdenas-Robledo et al.~\cite{ADRIANACARDENASROBLEDO2022101863} and extended it with additional dimensions to describe robots (the type, the range of functions, whether Wizard-of-Oz (WoZ) was used, and the way of interaction). Accordingly, the codes shown in Table \ref{tab:ff1_results_compact} were identified. All subcategories were created inductively, except those under \textit{role/context of XR}, which were adopted from~\cite{ADRIANACARDENASROBLEDO2022101863}. For \textbf{RQ2}, codes for data collection and analysis methods were developed following Döring \cite{doring2022forschungsmethoden}, which provides established guidance for operationalizing and categorizing research methods. All additional codes belonging to the research design were based on Creswell \cite{creswell2014researchdesign}, a widely used methodological guide to research design and its core design types, as can be seen in Table \ref{tab:methoden_ff1}. For \textbf{RQ3}, all subcategories under \textit{challenges} and \textit{future work} were created and categorized inductively (see Table \ref{tab:challenges_futurework_xr_hri}). For \textbf{RQ4}, codes were developed deductively. Common demographic data such as discipline, status, gender, country, and age were collected for both researchers and participants. For researchers, this information was identified via search engines.


\begin{table}[htbp]
\centering
\scriptsize
\setlength{\tabcolsep}{2.5pt}
\renewcommand{\arraystretch}{0.93}

\caption{Overview of codes and results from RQ1}
\label{tab:ff1_results_compact}

\begin{tabularx}{\linewidth}{@{} l l c >{\raggedright\arraybackslash}X @{}}
\toprule
\textbf{Code} & \textbf{Subcategory} & \textbf{Count} & \textbf{Paper} \\
\midrule

\multicolumn{4}{@{}l}{\textbf{Role/Context of XR}}\\
\addlinespace[1pt]
 & Evaluation \& Testing & 27 &
 \cite{Arpaia2022WearableBCI,Bjorling2022IAmTheRobot,Brown2023MixedRealityDeictic,Chenlin2023PerspectiveTakingProsocial,Gao2019MetaRLTrustHRI,Groechel2021KinestheticCuriosity,Groechel2019ExpressiveMRAms,Helgert2024UnderstandableTransparency,Karakosta2023SAROptimizingAR,Kim2019CollaborativeVRGameTeens,Lee2017AppearanceTherapyEngagement,Li2018PostureEmbodimentSocialDistance,Mahajan2020UsabilityMetricsMR,Mizuchi2022VRGUIBehaviorCollection,Mueller2017RoboticWorkmate,Nertinger2024RobotPostureSafety,Nigro2024InteractiveARProxemics,Peters2018SocialDistancesMR,Petrak2019ProxemicAwarenessFirstImpressions,Pilacinski2023RobotEyesTrustPerformance,Pozharliev2021AttachmentStyleResponses,Sadka2020VRSimulationNonHumanoid,Schulten2025MergingRealities,Strassmann2024DontJudgeBook,Wang2019EmbodimentSubstrateDecisionMaking,Zhang2025PredictingPerceptionsNavigation,Zhiyu2024ArchitecturalRoboticsRestorative} \\
 & Simulation & 24 &
 \cite{Bjorling2022IAmTheRobot,Brown2023MixedRealityDeictic,Chenlin2023PerspectiveTakingProsocial,Fujii2020CoEatingMR,Gao2019MetaRLTrustHRI,Groechel2021KinestheticCuriosity,Groechel2019ExpressiveMRAms,Groechel2023MoveToCodeAR,Helgert2024UnderstandableTransparency,Mizuchi2022VRGUIBehaviorCollection,Mueller2017RoboticWorkmate,Nertinger2024RobotPostureSafety,Nigro2024InteractiveARProxemics,Peters2018SocialDistancesMR,Petrak2019ProxemicAwarenessFirstImpressions,Pilacinski2023RobotEyesTrustPerformance,Pozharliev2021AttachmentStyleResponses,Sadka2020VRSimulationNonHumanoid,Schulten2025MergingRealities,Strassmann2024DontJudgeBook,Wang2019EmbodimentSubstrateDecisionMaking,Wijnen2020HRIUserStudiesVR,Ye2022RCareWorld,Zhiyu2024ArchitecturalRoboticsRestorative} \\
 & Training \& Education & 8 &
 \cite{Arpaia2022WearableBCI,Bjorling2022IAmTheRobot,Groechel2021KinestheticCuriosity,Groechel2023MoveToCodeAR,Karakosta2023SAROptimizingAR,Mahajan2020UsabilityMetricsMR,Mizuchi2022VRGUIBehaviorCollection,Wijnen2020HRIUserStudiesVR} \\
 & Prototyping \& Design & 8 &
 \cite{Bjorling2022IAmTheRobot,Gao2019MetaRLTrustHRI,Helgert2025aVRStudyTool,Kim2019CollaborativeVRGameTeens,Mueller2017RoboticWorkmate,Nigro2024InteractiveARProxemics,Petrak2019ProxemicAwarenessFirstImpressions,Ye2022RCareWorld} \\
 & Collaboration & 4 &
 \cite{Bjorling2022IAmTheRobot,Brown2023MixedRealityDeictic,Fan2022SARConnect,Kim2019CollaborativeVRGameTeens} \\
\midrule

\multicolumn{4}{@{}l}{\textbf{Field of Application}}\\
\addlinespace[1pt]
 & Cross-Domain & 12 &
 \cite{Brown2023MixedRealityDeictic,Chenlin2023PerspectiveTakingProsocial,Gao2019MetaRLTrustHRI,Helgert2025aVRStudyTool,Li2018PostureEmbodimentSocialDistance,Mizuchi2022VRGUIBehaviorCollection,Peters2018SocialDistancesMR,Petrak2019ProxemicAwarenessFirstImpressions,Pilacinski2023RobotEyesTrustPerformance,Sadka2020VRSimulationNonHumanoid,Wang2019EmbodimentSubstrateDecisionMaking,Zhang2025PredictingPerceptionsNavigation} \\
 & Education & 7 &
 \cite{Bjorling2022IAmTheRobot,Groechel2021KinestheticCuriosity,Groechel2023MoveToCodeAR,Groechel2019ExpressiveMRAms,Karakosta2023SAROptimizingAR,Kim2019CollaborativeVRGameTeens,Mahajan2020UsabilityMetricsMR} \\
 & Healthcare & 7 &
 \cite{Arpaia2022WearableBCI,Fan2022SARConnect,Lee2017AppearanceTherapyEngagement,Nertinger2024RobotPostureSafety,Nigro2024InteractiveARProxemics,Ye2022RCareWorld,Zhiyu2024ArchitecturalRoboticsRestorative} \\
 & Public Space & 4 &
 \cite{Helgert2024UnderstandableTransparency,Schulten2025MergingRealities,Strassmann2024DontJudgeBook,Wijnen2020HRIUserStudiesVR} \\
 & Industry & 1 & \cite{Mueller2017RoboticWorkmate} \\
 & Medical Facilities & 1 & \cite{Pozharliev2021AttachmentStyleResponses} \\
\midrule

\multicolumn{4}{@{}l}{\textbf{XR Technology}}\\
\addlinespace[1pt]
 & VR & 20 &
 \cite{Bjorling2022IAmTheRobot,Chenlin2023PerspectiveTakingProsocial,Fan2022SARConnect,Helgert2025aVRStudyTool,Helgert2024UnderstandableTransparency,Kim2019CollaborativeVRGameTeens,Lee2017AppearanceTherapyEngagement,Mizuchi2022VRGUIBehaviorCollection,Mueller2017RoboticWorkmate,Nertinger2024RobotPostureSafety,Petrak2019ProxemicAwarenessFirstImpressions,Pilacinski2023RobotEyesTrustPerformance,Pozharliev2021AttachmentStyleResponses,Sadka2020VRSimulationNonHumanoid,Strassmann2024DontJudgeBook,Wijnen2020HRIUserStudiesVR,Ye2022RCareWorld,Zhang2025PredictingPerceptionsNavigation,Zhiyu2024ArchitecturalRoboticsRestorative} \\
 & MR & 9 &
 \cite{Brown2023MixedRealityDeictic,Fujii2020CoEatingMR,Gao2019MetaRLTrustHRI,Groechel2021KinestheticCuriosity,Groechel2019ExpressiveMRAms,Li2018PostureEmbodimentSocialDistance,Mahajan2020UsabilityMetricsMR,Peters2018SocialDistancesMR,Schulten2025MergingRealities} \\
 & AR & 5 &
 \cite{Arpaia2022WearableBCI,Groechel2023MoveToCodeAR,Karakosta2023SAROptimizingAR,Nigro2024InteractiveARProxemics,Wang2019EmbodimentSubstrateDecisionMaking} \\
\midrule

\multicolumn{4}{@{}l}{\textbf{XR Hardware}}\\
\addlinespace[1pt]
 & MS HoloLens & 8 &
 \cite{Brown2023MixedRealityDeictic,Gao2019MetaRLTrustHRI,Groechel2021KinestheticCuriosity,Groechel2019ExpressiveMRAms,Li2018PostureEmbodimentSocialDistance,Mahajan2020UsabilityMetricsMR,Peters2018SocialDistancesMR,Wang2019EmbodimentSubstrateDecisionMaking} \\
 & HTC Vive & 7 &
 \cite{Bjorling2022IAmTheRobot,Nertinger2024RobotPostureSafety,Petrak2019ProxemicAwarenessFirstImpressions,Pilacinski2023RobotEyesTrustPerformance,Pozharliev2021AttachmentStyleResponses,Wijnen2020HRIUserStudiesVR,Zhang2025PredictingPerceptionsNavigation} \\
 & Meta Quest & 4 &
 \cite{Chenlin2023PerspectiveTakingProsocial,Schulten2025MergingRealities,Helgert2025aVRStudyTool,Strassmann2024DontJudgeBook} \\
 & Others & 5 &
 \cite{Mahajan2020UsabilityMetricsMR,Lee2017AppearanceTherapyEngagement,Pozharliev2021AttachmentStyleResponses,Arpaia2022WearableBCI,Pilacinski2023RobotEyesTrustPerformance} \\
 & Not specified & 12 &
 \cite{Fan2022SARConnect,Helgert2024UnderstandableTransparency,Kim2019CollaborativeVRGameTeens,Lee2017AppearanceTherapyEngagement,Mizuchi2022VRGUIBehaviorCollection,Mueller2017RoboticWorkmate,Sadka2020VRSimulationNonHumanoid,Ye2022RCareWorld,Zhiyu2024ArchitecturalRoboticsRestorative,Groechel2023MoveToCodeAR,Karakosta2023SAROptimizingAR,Nigro2024InteractiveARProxemics} \\
\midrule

\multicolumn{4}{@{}l}{\textbf{XR Software}}\\
\addlinespace[1pt]
 & Unity & 19 &
 \cite{Brown2023MixedRealityDeictic,Gao2019MetaRLTrustHRI,Groechel2019ExpressiveMRAms,Li2018PostureEmbodimentSocialDistance,Mahajan2020UsabilityMetricsMR,Peters2018SocialDistancesMR,Schulten2025MergingRealities,Bjorling2022IAmTheRobot,Chenlin2023PerspectiveTakingProsocial,Helgert2025aVRStudyTool,Helgert2024UnderstandableTransparency,Mueller2017RoboticWorkmate,Nertinger2024RobotPostureSafety,Pilacinski2023RobotEyesTrustPerformance,Pozharliev2021AttachmentStyleResponses,Strassmann2024DontJudgeBook,Ye2022RCareWorld,Zhang2025PredictingPerceptionsNavigation,Nigro2024InteractiveARProxemics} \\
 & Custom-built & 6 &
 \cite{Groechel2021KinestheticCuriosity,Fan2022SARConnect,Kim2019CollaborativeVRGameTeens,Lee2017AppearanceTherapyEngagement,Zhang2025PredictingPerceptionsNavigation,Groechel2023MoveToCodeAR} \\
 & ROS & 5 &
 \cite{Groechel2021KinestheticCuriosity,Groechel2019ExpressiveMRAms,Pilacinski2023RobotEyesTrustPerformance,Ye2022RCareWorld,Zhang2025PredictingPerceptionsNavigation} \\
 & Unreal & 1 & \cite{Wijnen2020HRIUserStudiesVR} \\
 & Not specified & 5 &
 \cite{Mizuchi2022VRGUIBehaviorCollection,Petrak2019ProxemicAwarenessFirstImpressions,Sadka2020VRSimulationNonHumanoid,Zhiyu2024ArchitecturalRoboticsRestorative,Wang2019EmbodimentSubstrateDecisionMaking} \\
\midrule

\multicolumn{4}{@{}l}{\textbf{Robots}}\\
\addlinespace[1pt]
 & Pepper & 8 &
 \cite{Gao2019MetaRLTrustHRI,Helgert2025aVRStudyTool,Helgert2024UnderstandableTransparency,Li2018PostureEmbodimentSocialDistance,Peters2018SocialDistancesMR,Pozharliev2021AttachmentStyleResponses,Schulten2025MergingRealities,Strassmann2024DontJudgeBook} \\
 & NAO & 6 &
 \cite{Fan2022SARConnect,Fujii2020CoEatingMR,Karakosta2023SAROptimizingAR,Li2018PostureEmbodimentSocialDistance,Wang2019EmbodimentSubstrateDecisionMaking,Ye2022RCareWorld} \\
 & Custom-built & 6 &
 \cite{Bjorling2022IAmTheRobot,Kim2019CollaborativeVRGameTeens,Mueller2017RoboticWorkmate,Petrak2019ProxemicAwarenessFirstImpressions,Sadka2020VRSimulationNonHumanoid,Zhiyu2024ArchitecturalRoboticsRestorative} \\
 & Kuri & 4 &
 \cite{Groechel2019ExpressiveMRAms,Groechel2021KinestheticCuriosity,Groechel2023MoveToCodeAR,Nigro2024InteractiveARProxemics} \\
 & Others & 7 &
 \cite{Zhang2025PredictingPerceptionsNavigation,Wijnen2020HRIUserStudiesVR,Pilacinski2023RobotEyesTrustPerformance,Nertinger2024RobotPostureSafety,Brown2023MixedRealityDeictic,Arpaia2022WearableBCI} \\
 & Not specified & 3 &
 \cite{Chenlin2023PerspectiveTakingProsocial,Mahajan2020UsabilityMetricsMR,Mizuchi2022VRGUIBehaviorCollection} \\
\midrule

\multicolumn{4}{@{}l}{\textbf{Wizard of Oz}}\\
\addlinespace[1pt]
 & Yes & 12 &
 \cite{Bjorling2022IAmTheRobot,Helgert2025aVRStudyTool,Helgert2024UnderstandableTransparency,Karakosta2023SAROptimizingAR,Li2018PostureEmbodimentSocialDistance,Mizuchi2022VRGUIBehaviorCollection,Mueller2017RoboticWorkmate,Sadka2020VRSimulationNonHumanoid,Strassmann2024DontJudgeBook,Wang2019EmbodimentSubstrateDecisionMaking,Wijnen2020HRIUserStudiesVR,Zhiyu2024ArchitecturalRoboticsRestorative} \\
 & No & 21 &
 \cite{Arpaia2022WearableBCI,Brown2023MixedRealityDeictic,Chenlin2023PerspectiveTakingProsocial,Fujii2020CoEatingMR,Gao2019MetaRLTrustHRI,Groechel2021KinestheticCuriosity,Groechel2019ExpressiveMRAms,Groechel2023MoveToCodeAR,Kim2019CollaborativeVRGameTeens,Lee2017AppearanceTherapyEngagement,Mahajan2020UsabilityMetricsMR,Nertinger2024RobotPostureSafety,Nigro2024InteractiveARProxemics,Peters2018SocialDistancesMR,Petrak2019ProxemicAwarenessFirstImpressions,Pilacinski2023RobotEyesTrustPerformance,Pozharliev2021AttachmentStyleResponses,Schulten2025MergingRealities,Zhang2025PredictingPerceptionsNavigation,Ye2022RCareWorld,Fan2022SARConnect} \\
\midrule

\multicolumn{4}{@{}l}{\textbf{Robot Functionality}}\\
\addlinespace[1pt]
 & Speech & 20 &
 \cite{Fan2022SARConnect,Fujii2020CoEatingMR,Gao2019MetaRLTrustHRI,Groechel2021KinestheticCuriosity,Groechel2023MoveToCodeAR,Helgert2025aVRStudyTool,Helgert2024UnderstandableTransparency,Karakosta2023SAROptimizingAR,Lee2017AppearanceTherapyEngagement,Li2018PostureEmbodimentSocialDistance,Mahajan2020UsabilityMetricsMR,Mizuchi2022VRGUIBehaviorCollection,Mueller2017RoboticWorkmate,Pozharliev2021AttachmentStyleResponses,Schulten2025MergingRealities,Strassmann2024DontJudgeBook,Wang2019EmbodimentSubstrateDecisionMaking,Wijnen2020HRIUserStudiesVR,Ye2022RCareWorld,Zhiyu2024ArchitecturalRoboticsRestorative} \\
 & Gestures & 15 &
 \cite{Brown2023MixedRealityDeictic,Fan2022SARConnect,Fujii2020CoEatingMR,Groechel2021KinestheticCuriosity,Groechel2023MoveToCodeAR,Helgert2025aVRStudyTool,Helgert2024UnderstandableTransparency,Karakosta2023SAROptimizingAR,Lee2017AppearanceTherapyEngagement,Mizuchi2022VRGUIBehaviorCollection,Sadka2020VRSimulationNonHumanoid,Schulten2025MergingRealities,Strassmann2024DontJudgeBook,Wang2019EmbodimentSubstrateDecisionMaking,Ye2022RCareWorld} \\
 & Navigation & 13 &
 \cite{Arpaia2022WearableBCI,Chenlin2023PerspectiveTakingProsocial,Groechel2023MoveToCodeAR,Helgert2025aVRStudyTool,Nertinger2024RobotPostureSafety,Nigro2024InteractiveARProxemics,Petrak2019ProxemicAwarenessFirstImpressions,Pilacinski2023RobotEyesTrustPerformance,Sadka2020VRSimulationNonHumanoid,Schulten2025MergingRealities,Wijnen2020HRIUserStudiesVR,Zhang2025PredictingPerceptionsNavigation,Zhiyu2024ArchitecturalRoboticsRestorative} \\
 & Others & 4 &
 \cite{Helgert2025aVRStudyTool,Li2018PostureEmbodimentSocialDistance,Helgert2024UnderstandableTransparency,Chenlin2023PerspectiveTakingProsocial} \\
\midrule

\multicolumn{4}{@{}l}{\textbf{Type of interaction}}\\
\addlinespace[1pt]
 & No interaction & 15 &
 \cite{Bjorling2022IAmTheRobot,Brown2023MixedRealityDeictic,Fan2022SARConnect,Fujii2020CoEatingMR,Groechel2021KinestheticCuriosity,Groechel2023MoveToCodeAR,Groechel2019ExpressiveMRAms,Kim2019CollaborativeVRGameTeens,Mahajan2020UsabilityMetricsMR,Nertinger2024RobotPostureSafety,Peters2018SocialDistancesMR,Petrak2019ProxemicAwarenessFirstImpressions,Sadka2020VRSimulationNonHumanoid,Wijnen2020HRIUserStudiesVR,Zhang2025PredictingPerceptionsNavigation} \\
 & Voice Commands & 10 &
 \cite{Helgert2024UnderstandableTransparency,Lee2017AppearanceTherapyEngagement,Li2018PostureEmbodimentSocialDistance,Mizuchi2022VRGUIBehaviorCollection,Mueller2017RoboticWorkmate,Wang2019EmbodimentSubstrateDecisionMaking,Zhiyu2024ArchitecturalRoboticsRestorative,Gao2019MetaRLTrustHRI,Karakosta2023SAROptimizingAR,Pozharliev2021AttachmentStyleResponses} \\
 & Object Handling & 3 & \cite{Pilacinski2023RobotEyesTrustPerformance,Strassmann2024DontJudgeBook,Ye2022RCareWorld} \\
 & Tablet & 2 & \cite{Nigro2024InteractiveARProxemics,Schulten2025MergingRealities} \\
 & Human Moves Robot & 2 & \cite{Arpaia2022WearableBCI,Chenlin2023PerspectiveTakingProsocial} \\
\bottomrule
\end{tabularx}
\end{table}

\section{Results}

\subsection{RQ1: Roles and Contexts of Extended Reality}

The \textbf{roles and contexts} of XR in the reviewed studies are dominated by \textit{evaluation \& testing} (\textit{n=27}) and \textit{simulation} (\textit{n=24}). In about three-quarters of the papers, XR functions as a controllable laboratory environment for risk-free variation and measurement of robot concepts. Less common are roles such as \textit{training \& education} (\textit{n=8}) or \textit{prototyping \& design} (\textit{n=8}), while \textit{robot–robot collaboration} (\textit{n=4}) remains rare. XR is thus used mainly for observation rather than active co-creative interaction. Regarding the \textbf{field of application}, more than one third of studies examine fundamental HRI phenomena without a specific context (\textit{n=12}). Among applied scenarios, \textit{education} (\textit{n=7}) is most prominent, followed by \textit{healthcare} (\textit{n=7}), where XR supports safe, repeatable simulations for learning and rehabilitation. \textit{Public and commercial settings} (\textit{n=4}) receive comparatively little attention. In terms of \textbf{XR technologies}, \textit{VR clearly dominates} (\textit{n=20}) due to technical maturity and high experimental control. \textit{MR is emerging} (\textit{n=9}) but remains limited by development complexity ~\cite{buildings13040872}. \textit{AR plays only a minor role} (\textit{n=5}). Common devices include HTC Vive/Pro (\textit{n=7}) and Microsoft HoloLens 2 (\textit{n=8}), with Unity as the prevalent development framework (\textit{n=19}), whereby some studies fail to report hardware/software details. Concerning the virtual \textbf{robot platforms}, \textit{Pepper} (\textit{n=8}) and \textit{NAO} (\textit{n=6}) are most frequently used, while a few studies leverage XR to implement \textit{entirely virtual robots} (\textit{n=6}). Over one third rely on \textit{Wizard-of-Oz} (\textit{n=23}) control, mainly to simulate naturalistic speech. \textbf{Robot functionality} is dominated by \textit{speech (\textit{n=20}), gestures (\textit{n=15}), and navigation (\textit{n=13})}, typically through pre-defined utterances that ensure control but reduce ecological validity ~\cite{Parsons2015}. \textit{Visual cues} or \textit{tablet} (\textit{n=4}) interactions are less common, and some robots serve merely as passive stimuli. With respect to \textbf{interaction types}, nearly half of the studies involve \textit{no participant–robot interaction} (\textit{n=15}) at all. When interaction is allowed, it usually consists of \textit{short, pre-scripted speech or tablet commands} (\textit{n=10}). Only a small number of studies support \textit{dynamic speech interaction}, where participants could freely interact with the robot without having to adhere to predefined voice commands, mostly handled via Wizard-of-Oz mechanisms.

\subsection{RQ2: Data \& Analysis Collection and Research Design}

Regarding \textbf{data collection methods}, \textit{questionnaires (\textit{n=29}) and observations (\textit{n=25})} are used most frequently, while \textit{interviews} (\textit{n=10}) appear only occasionally. \textit{Physiological measurements} (\textit{n=7}) remain rare despite the availability of integrated headset sensors. \textit{Real-time metrics} such as eye tracking, pupillometry, or cognitive workload (e.g., via HP Reverb G2 Omnicept) are largely underutilized, and standardized psychological tests (e.g., Implicit Association Test) appear only in isolated cases (\textit{n=3}). As a result, social XR-HRI research continues to rely mainly on subjective self-reports and manual behavioral coding. Concerning \textbf{data analysis methods}, all studies report \textit{descriptive statistics} (\textit{n=33}), and most employ \textit{inferential analyses} (\textit{n=27}). \textit{Comparative statistics} (\textit{n=16}) appear in nearly half of the papers, while \textit{correlation (\textit{n=10}) and qualitative analyses (\textit{n=9})} are found primarily in mixed-methods designs. \textit{Applied research dominates} (\textit{n=22}), making up two-thirds of the studies; all works are \textit{empirical} (\textit{n=33}) and analyze \textit{primary data} (\textit{n=33}). Most follow an \textit{explanatory} (\textit{n=21}) approach, though \textit{exploratory} (\textit{n=15}) components reflect the field’s novelty. Experiments are predominantly conducted in \textit{laboratory settings} (\textit{n=29}), which aligns with the XR focus. The \textit{few field studies} (\textit{n=4}) identified all involve AR. Regarding measurement timing, \textit{repeated-measures designs} (\textit{n=14}) are most common, whereas cross-sectional (\textit{n=10}) and single-subject (\textit{n=9}) approaches occur less frequently.

\begin{table}[htbp]
\centering
\scriptsize
\setlength{\tabcolsep}{2.5pt}
\renewcommand{\arraystretch}{0.93}

\caption{Overview of codes and results from RQ2}
\label{tab:methoden_ff1}

\begin{tabularx}{\linewidth}{@{} l l c >{\raggedright\arraybackslash}X @{}}
\toprule
\textbf{Code} & \textbf{Subcategory} & \textbf{Count} & \textbf{Paper} \\
\midrule

\multicolumn{4}{@{}l}{\textbf{Data Collection Methods}}\\
\addlinespace[1pt]
 & Questionnaire & 29 &
 \cite{Bjorling2022IAmTheRobot,Brown2023MixedRealityDeictic,Chenlin2023PerspectiveTakingProsocial,Fujii2020CoEatingMR,Gao2019MetaRLTrustHRI,Groechel2019ExpressiveMRAms,Groechel2021KinestheticCuriosity,Helgert2024UnderstandableTransparency,Helgert2025aVRStudyTool,Karakosta2023SAROptimizingAR,Kim2019CollaborativeVRGameTeens,Lee2017AppearanceTherapyEngagement,Li2018PostureEmbodimentSocialDistance,Mahajan2020UsabilityMetricsMR,Mizuchi2022VRGUIBehaviorCollection,Mueller2017RoboticWorkmate,Nertinger2024RobotPostureSafety,Nigro2024InteractiveARProxemics,Peters2018SocialDistancesMR,Petrak2019ProxemicAwarenessFirstImpressions,Pilacinski2023RobotEyesTrustPerformance,Pozharliev2021AttachmentStyleResponses,Sadka2020VRSimulationNonHumanoid,Schulten2025MergingRealities,Strassmann2024DontJudgeBook,Wang2019EmbodimentSubstrateDecisionMaking,Wijnen2020HRIUserStudiesVR,Zhang2025PredictingPerceptionsNavigation,Zhiyu2024ArchitecturalRoboticsRestorative} \\
 & Observation & 25 &
 \cite{Arpaia2022WearableBCI,Bjorling2022IAmTheRobot,Brown2023MixedRealityDeictic,Chenlin2023PerspectiveTakingProsocial,Fan2022SARConnect,Fujii2020CoEatingMR,Gao2019MetaRLTrustHRI,Groechel2019ExpressiveMRAms,Groechel2021KinestheticCuriosity,Groechel2023MoveToCodeAR,Helgert2025aVRStudyTool,Karakosta2023SAROptimizingAR,Kim2019CollaborativeVRGameTeens,Mahajan2020UsabilityMetricsMR,Mizuchi2022VRGUIBehaviorCollection,Mueller2017RoboticWorkmate,Nertinger2024RobotPostureSafety,Nigro2024InteractiveARProxemics,Peters2018SocialDistancesMR,Pilacinski2023RobotEyesTrustPerformance,Wang2019EmbodimentSubstrateDecisionMaking,Wijnen2020HRIUserStudiesVR,Ye2022RCareWorld,Zhang2025PredictingPerceptionsNavigation,Zhiyu2024ArchitecturalRoboticsRestorative} \\
 & Interview & 10 &
 \cite{Bjorling2022IAmTheRobot,Chenlin2023PerspectiveTakingProsocial,Groechel2021KinestheticCuriosity,Helgert2025aVRStudyTool,Karakosta2023SAROptimizingAR,Li2018PostureEmbodimentSocialDistance,Nigro2024InteractiveARProxemics,Schulten2025MergingRealities,Ye2022RCareWorld,Zhiyu2024ArchitecturalRoboticsRestorative} \\
 & Physiological Measurements & 7 &
 \cite{Fan2022SARConnect,Lee2017AppearanceTherapyEngagement,Mahajan2020UsabilityMetricsMR,Pilacinski2023RobotEyesTrustPerformance,Pozharliev2021AttachmentStyleResponses,Ye2022RCareWorld,Zhang2025PredictingPerceptionsNavigation} \\
 & Psychological Test & 3 &
 \cite{Arpaia2022WearableBCI,Fan2022SARConnect,Groechel2023MoveToCodeAR} \\
 & Think Aloud & 1 & \cite{Helgert2025aVRStudyTool} \\
\midrule

\multicolumn{4}{@{}l}{\textbf{Data Analysis Methods}}\\
\addlinespace[1pt]
 & Descriptive Statistics & 33 &
 \cite{Arpaia2022WearableBCI,Bjorling2022IAmTheRobot,Brown2023MixedRealityDeictic,Chenlin2023PerspectiveTakingProsocial,Fan2022SARConnect,Fujii2020CoEatingMR,Gao2019MetaRLTrustHRI,Groechel2019ExpressiveMRAms,Groechel2021KinestheticCuriosity,Groechel2023MoveToCodeAR,Helgert2024UnderstandableTransparency,Helgert2025aVRStudyTool,Karakosta2023SAROptimizingAR,Kim2019CollaborativeVRGameTeens,Lee2017AppearanceTherapyEngagement,Li2018PostureEmbodimentSocialDistance,Mahajan2020UsabilityMetricsMR,Mizuchi2022VRGUIBehaviorCollection,Mueller2017RoboticWorkmate,Nertinger2024RobotPostureSafety,Nigro2024InteractiveARProxemics,Peters2018SocialDistancesMR,Petrak2019ProxemicAwarenessFirstImpressions,Pilacinski2023RobotEyesTrustPerformance,Pozharliev2021AttachmentStyleResponses,Sadka2020VRSimulationNonHumanoid,Schulten2025MergingRealities,Strassmann2024DontJudgeBook,Wang2019EmbodimentSubstrateDecisionMaking,Wijnen2020HRIUserStudiesVR,Ye2022RCareWorld,Zhang2025PredictingPerceptionsNavigation,Zhiyu2024ArchitecturalRoboticsRestorative} \\
 & Inferential Statistics & 27 &
 \cite{Brown2023MixedRealityDeictic,Chenlin2023PerspectiveTakingProsocial,Fan2022SARConnect,Fujii2020CoEatingMR,Gao2019MetaRLTrustHRI,Groechel2019ExpressiveMRAms,Groechel2023MoveToCodeAR,Helgert2024UnderstandableTransparency,Karakosta2023SAROptimizingAR,Lee2017AppearanceTherapyEngagement,Li2018PostureEmbodimentSocialDistance,Mahajan2020UsabilityMetricsMR,Mizuchi2022VRGUIBehaviorCollection,Mueller2017RoboticWorkmate,Nertinger2024RobotPostureSafety,Nigro2024InteractiveARProxemics,Peters2018SocialDistancesMR,Petrak2019ProxemicAwarenessFirstImpressions,Pilacinski2023RobotEyesTrustPerformance,Pozharliev2021AttachmentStyleResponses,Sadka2020VRSimulationNonHumanoid,Schulten2025MergingRealities,Strassmann2024DontJudgeBook,Wang2019EmbodimentSubstrateDecisionMaking,Wijnen2020HRIUserStudiesVR,Zhang2025PredictingPerceptionsNavigation,Zhiyu2024ArchitecturalRoboticsRestorative} \\
 & Comparative Analyses & 16 &
 \cite{Brown2023MixedRealityDeictic,Chenlin2023PerspectiveTakingProsocial,Fan2022SARConnect,Groechel2019ExpressiveMRAms,Helgert2024UnderstandableTransparency,Lee2017AppearanceTherapyEngagement,Li2018PostureEmbodimentSocialDistance,Mahajan2020UsabilityMetricsMR,Mizuchi2022VRGUIBehaviorCollection,Mueller2017RoboticWorkmate,Nigro2024InteractiveARProxemics,Petrak2019ProxemicAwarenessFirstImpressions,Pilacinski2023RobotEyesTrustPerformance,Strassmann2024DontJudgeBook,Wijnen2020HRIUserStudiesVR,Zhang2025PredictingPerceptionsNavigation} \\
 & Correlation Analyses & 10 &
 \cite{Groechel2019ExpressiveMRAms,Helgert2024UnderstandableTransparency,Mahajan2020UsabilityMetricsMR,Mizuchi2022VRGUIBehaviorCollection,Nigro2024InteractiveARProxemics,Pilacinski2023RobotEyesTrustPerformance,Pozharliev2021AttachmentStyleResponses,Schulten2025MergingRealities,Wijnen2020HRIUserStudiesVR,Zhang2025PredictingPerceptionsNavigation} \\
 & Qualitative Content Analyses & 9 &
 \cite{Bjorling2022IAmTheRobot,Groechel2019ExpressiveMRAms,Groechel2023MoveToCodeAR,Helgert2024UnderstandableTransparency,Karakosta2023SAROptimizingAR,Kim2019CollaborativeVRGameTeens,Li2018PostureEmbodimentSocialDistance,Schulten2025MergingRealities,Zhiyu2024ArchitecturalRoboticsRestorative} \\
\midrule

\multicolumn{4}{@{}l}{\textbf{Theoretical Approach}}\\
\addlinespace[1pt]
 & Quantitative Study & 17 &
 \cite{Arpaia2022WearableBCI,Brown2023MixedRealityDeictic,Chenlin2023PerspectiveTakingProsocial,Fan2022SARConnect,Gao2019MetaRLTrustHRI,Groechel2021KinestheticCuriosity,Lee2017AppearanceTherapyEngagement,Mizuchi2022VRGUIBehaviorCollection,Nigro2024InteractiveARProxemics,Peters2018SocialDistancesMR,Petrak2019ProxemicAwarenessFirstImpressions,Pilacinski2023RobotEyesTrustPerformance,Pozharliev2021AttachmentStyleResponses,Sadka2020VRSimulationNonHumanoid,Strassmann2024DontJudgeBook,Wang2019EmbodimentSubstrateDecisionMaking,Wijnen2020HRIUserStudiesVR} \\
 & Mixed-Methods & 13 &
 \cite{Bjorling2022IAmTheRobot,Fujii2020CoEatingMR,Groechel2023MoveToCodeAR,Groechel2019ExpressiveMRAms,Helgert2024UnderstandableTransparency,Karakosta2023SAROptimizingAR,Li2018PostureEmbodimentSocialDistance,Mahajan2020UsabilityMetricsMR,Mueller2017RoboticWorkmate,Nertinger2024RobotPostureSafety,Schulten2025MergingRealities,Zhang2025PredictingPerceptionsNavigation,Zhiyu2024ArchitecturalRoboticsRestorative} \\
 & Qualitative Study & 3 &
 \cite{Helgert2025aVRStudyTool,Kim2019CollaborativeVRGameTeens,Ye2022RCareWorld} \\
\midrule

\multicolumn{4}{@{}l}{\textbf{Objective of the Study}}\\
\addlinespace[1pt]
 & Applied Research & 22 &
 \cite{Arpaia2022WearableBCI,Brown2023MixedRealityDeictic,Fan2022SARConnect,Fujii2020CoEatingMR,Gao2019MetaRLTrustHRI,Groechel2019ExpressiveMRAms,Groechel2021KinestheticCuriosity,Groechel2023MoveToCodeAR,Helgert2024UnderstandableTransparency,Helgert2025aVRStudyTool,Karakosta2023SAROptimizingAR,Kim2019CollaborativeVRGameTeens,Lee2017AppearanceTherapyEngagement,Mahajan2020UsabilityMetricsMR,Mizuchi2022VRGUIBehaviorCollection,Mueller2017RoboticWorkmate,Pozharliev2021AttachmentStyleResponses,Schulten2025MergingRealities,Wijnen2020HRIUserStudiesVR,Ye2022RCareWorld,Zhang2025PredictingPerceptionsNavigation,Zhiyu2024ArchitecturalRoboticsRestorative} \\
 & Basic Research & 11 &
 \cite{Bjorling2022IAmTheRobot,Chenlin2023PerspectiveTakingProsocial,Li2018PostureEmbodimentSocialDistance,Nertinger2024RobotPostureSafety,Nigro2024InteractiveARProxemics,Peters2018SocialDistancesMR,Petrak2019ProxemicAwarenessFirstImpressions,Pilacinski2023RobotEyesTrustPerformance,Sadka2020VRSimulationNonHumanoid,Strassmann2024DontJudgeBook,Wang2019EmbodimentSubstrateDecisionMaking} \\
\midrule

\multicolumn{4}{@{}l}{\textbf{Subject of the Study}}\\
\addlinespace[1pt]
 & Empirical & 33 &
 \cite{Arpaia2022WearableBCI,Bjorling2022IAmTheRobot,Brown2023MixedRealityDeictic,Chenlin2023PerspectiveTakingProsocial,Fan2022SARConnect,Fujii2020CoEatingMR,Gao2019MetaRLTrustHRI,Groechel2019ExpressiveMRAms,Groechel2021KinestheticCuriosity,Groechel2023MoveToCodeAR,Helgert2024UnderstandableTransparency,Helgert2025aVRStudyTool,Karakosta2023SAROptimizingAR,Kim2019CollaborativeVRGameTeens,Lee2017AppearanceTherapyEngagement,Li2018PostureEmbodimentSocialDistance,Mahajan2020UsabilityMetricsMR,Mizuchi2022VRGUIBehaviorCollection,Mueller2017RoboticWorkmate,Nertinger2024RobotPostureSafety,Nigro2024InteractiveARProxemics,Peters2018SocialDistancesMR,Petrak2019ProxemicAwarenessFirstImpressions,Pilacinski2023RobotEyesTrustPerformance,Pozharliev2021AttachmentStyleResponses,Sadka2020VRSimulationNonHumanoid,Schulten2025MergingRealities,Strassmann2024DontJudgeBook,Wang2019EmbodimentSubstrateDecisionMaking,Wijnen2020HRIUserStudiesVR,Ye2022RCareWorld,Zhang2025PredictingPerceptionsNavigation,Zhiyu2024ArchitecturalRoboticsRestorative} \\
\midrule

\multicolumn{4}{@{}l}{\textbf{Data Basis}}\\
\addlinespace[1pt]
 & Primary Analysis & 33 &
 \cite{Arpaia2022WearableBCI,Bjorling2022IAmTheRobot,Brown2023MixedRealityDeictic,Chenlin2023PerspectiveTakingProsocial,Fan2022SARConnect,Fujii2020CoEatingMR,Gao2019MetaRLTrustHRI,Groechel2019ExpressiveMRAms,Groechel2021KinestheticCuriosity,Groechel2023MoveToCodeAR,Helgert2024UnderstandableTransparency,Helgert2025aVRStudyTool,Karakosta2023SAROptimizingAR,Kim2019CollaborativeVRGameTeens,Lee2017AppearanceTherapyEngagement,Li2018PostureEmbodimentSocialDistance,Mahajan2020UsabilityMetricsMR,Mizuchi2022VRGUIBehaviorCollection,Mueller2017RoboticWorkmate,Nertinger2024RobotPostureSafety,Nigro2024InteractiveARProxemics,Peters2018SocialDistancesMR,Petrak2019ProxemicAwarenessFirstImpressions,Pilacinski2023RobotEyesTrustPerformance,Pozharliev2021AttachmentStyleResponses,Sadka2020VRSimulationNonHumanoid,Schulten2025MergingRealities,Strassmann2024DontJudgeBook,Wang2019EmbodimentSubstrateDecisionMaking,Wijnen2020HRIUserStudiesVR,Ye2022RCareWorld,Zhang2025PredictingPerceptionsNavigation,Zhiyu2024ArchitecturalRoboticsRestorative} \\
\midrule

\multicolumn{4}{@{}l}{\textbf{Research Interest}}\\
\addlinespace[1pt]
 & Explanatory & 21 &
 \cite{Brown2023MixedRealityDeictic,Chenlin2023PerspectiveTakingProsocial,Fan2022SARConnect,Fujii2020CoEatingMR,Gao2019MetaRLTrustHRI,Groechel2021KinestheticCuriosity,Helgert2025aVRStudyTool,Karakosta2023SAROptimizingAR,Lee2017AppearanceTherapyEngagement,Nertinger2024RobotPostureSafety,Nigro2024InteractiveARProxemics,Peters2018SocialDistancesMR,Petrak2019ProxemicAwarenessFirstImpressions,Pilacinski2023RobotEyesTrustPerformance,Pozharliev2021AttachmentStyleResponses,Sadka2020VRSimulationNonHumanoid,Strassmann2024DontJudgeBook,Wang2019EmbodimentSubstrateDecisionMaking,Wijnen2020HRIUserStudiesVR,Zhang2025PredictingPerceptionsNavigation,Zhiyu2024ArchitecturalRoboticsRestorative} \\
 & Exploratory & 15 &
 \cite{Arpaia2022WearableBCI,Bjorling2022IAmTheRobot,Groechel2023MoveToCodeAR,Groechel2019ExpressiveMRAms,Helgert2024UnderstandableTransparency,Kim2019CollaborativeVRGameTeens,Li2018PostureEmbodimentSocialDistance,Mahajan2020UsabilityMetricsMR,Mizuchi2022VRGUIBehaviorCollection,Mueller2017RoboticWorkmate,Nigro2024InteractiveARProxemics,Schulten2025MergingRealities,Ye2022RCareWorld,Zhiyu2024ArchitecturalRoboticsRestorative} \\
 & Descriptive & 12 &
 \cite{Arpaia2022WearableBCI,Fujii2020CoEatingMR,Gao2019MetaRLTrustHRI,Groechel2019ExpressiveMRAms,Groechel2021KinestheticCuriosity,Groechel2023MoveToCodeAR,Helgert2024UnderstandableTransparency,Mizuchi2022VRGUIBehaviorCollection,Nertinger2024RobotPostureSafety,Peters2018SocialDistancesMR,Petrak2019ProxemicAwarenessFirstImpressions,Pilacinski2023RobotEyesTrustPerformance} \\
\midrule

\multicolumn{4}{@{}l}{\textbf{Formation of Study Groups}}\\
\addlinespace[1pt]
 & Experimental Study & 19 &
 \cite{Arpaia2022WearableBCI,Brown2023MixedRealityDeictic,Chenlin2023PerspectiveTakingProsocial,Fan2022SARConnect,Fujii2020CoEatingMR,Gao2019MetaRLTrustHRI,Groechel2019ExpressiveMRAms,Groechel2021KinestheticCuriosity,Lee2017AppearanceTherapyEngagement,Li2018PostureEmbodimentSocialDistance,Nertinger2024RobotPostureSafety,Nigro2024InteractiveARProxemics,Peters2018SocialDistancesMR,Petrak2019ProxemicAwarenessFirstImpressions,Pilacinski2023RobotEyesTrustPerformance,Pozharliev2021AttachmentStyleResponses,Strassmann2024DontJudgeBook,Wang2019EmbodimentSubstrateDecisionMaking,Zhang2025PredictingPerceptionsNavigation} \\
 & Non-Experimental Study & 8 &
 \cite{Bjorling2022IAmTheRobot,Helgert2025aVRStudyTool,Kim2019CollaborativeVRGameTeens,Mahajan2020UsabilityMetricsMR,Mizuchi2022VRGUIBehaviorCollection,Mueller2017RoboticWorkmate,Schulten2025MergingRealities,Ye2022RCareWorld} \\
 & Quasi-Experimental Study & 6 &
 \cite{Groechel2023MoveToCodeAR,Helgert2024UnderstandableTransparency,Karakosta2023SAROptimizingAR,Sadka2020VRSimulationNonHumanoid,Wijnen2020HRIUserStudiesVR,Zhiyu2024ArchitecturalRoboticsRestorative} \\
\midrule

\multicolumn{4}{@{}l}{\textbf{Study Location}}\\
\addlinespace[1pt]
 & Laboratory & 29 &
 \cite{Bjorling2022IAmTheRobot,Brown2023MixedRealityDeictic,Chenlin2023PerspectiveTakingProsocial,Fan2022SARConnect,Fujii2020CoEatingMR,Gao2019MetaRLTrustHRI,Groechel2019ExpressiveMRAms,Groechel2021KinestheticCuriosity,Helgert2024UnderstandableTransparency,Helgert2025aVRStudyTool,Kim2019CollaborativeVRGameTeens,Lee2017AppearanceTherapyEngagement,Li2018PostureEmbodimentSocialDistance,Mahajan2020UsabilityMetricsMR,Mizuchi2022VRGUIBehaviorCollection,Mueller2017RoboticWorkmate,Nertinger2024RobotPostureSafety,Nigro2024InteractiveARProxemics,Peters2018SocialDistancesMR,Petrak2019ProxemicAwarenessFirstImpressions,Pilacinski2023RobotEyesTrustPerformance,Pozharliev2021AttachmentStyleResponses,Sadka2020VRSimulationNonHumanoid,Strassmann2024DontJudgeBook,Wang2019EmbodimentSubstrateDecisionMaking,Wijnen2020HRIUserStudiesVR,Ye2022RCareWorld,Zhang2025PredictingPerceptionsNavigation,Zhiyu2024ArchitecturalRoboticsRestorative} \\
 & Field & 4 &
 \cite{Arpaia2022WearableBCI,Groechel2023MoveToCodeAR,Karakosta2023SAROptimizingAR,Schulten2025MergingRealities} \\
\midrule

\multicolumn{4}{@{}l}{\textbf{Number of Measurements}}\\
\addlinespace[1pt]
 & Repeated & 14 &
 \cite{Arpaia2022WearableBCI,Brown2023MixedRealityDeictic,Fujii2020CoEatingMR,Gao2019MetaRLTrustHRI,Groechel2021KinestheticCuriosity,Groechel2023MoveToCodeAR,Karakosta2023SAROptimizingAR,Li2018PostureEmbodimentSocialDistance,Nertinger2024RobotPostureSafety,Nigro2024InteractiveARProxemics,Peters2018SocialDistancesMR,Petrak2019ProxemicAwarenessFirstImpressions,Pilacinski2023RobotEyesTrustPerformance,Zhiyu2024ArchitecturalRoboticsRestorative} \\
 & Cross-Sectional & 10 &
 \cite{Groechel2019ExpressiveMRAms,Helgert2024UnderstandableTransparency,Helgert2025aVRStudyTool,Lee2017AppearanceTherapyEngagement,Mahajan2020UsabilityMetricsMR,Mizuchi2022VRGUIBehaviorCollection,Pozharliev2021AttachmentStyleResponses,Sadka2020VRSimulationNonHumanoid,Schulten2025MergingRealities,Ye2022RCareWorld} \\
 & Single & 9 &
 \cite{Bjorling2022IAmTheRobot,Chenlin2023PerspectiveTakingProsocial,Fan2022SARConnect,Kim2019CollaborativeVRGameTeens,Mueller2017RoboticWorkmate,Strassmann2024DontJudgeBook,Wang2019EmbodimentSubstrateDecisionMaking,Wijnen2020HRIUserStudiesVR,Zhang2025PredictingPerceptionsNavigation} \\
\bottomrule
\end{tabularx}
\end{table}

\subsection{RQ3: Current Challenges and Future Research}

The most frequently reported \textbf{challenges} concern \textbf{technical limitations and hardware} (\textit{n=21}). Restricted field of view, comfort issues, latency, and visual inconsistencies often disrupted immersion. Inaccurate position or depth tracking, network instability (e.g., UDP delays), and asynchronous timing between robot and display further reduced interaction quality. Software crashes, extensive calibration in WoZ setups, and the absence of haptic feedback or intuitive interfaces weakened usability. Overall, technology remains a major bottleneck, underscoring the need for robust hardware, stable networks, and XR-aware interaction design. Regarding \textbf{methodological limitations} (\textit{n=21}), many studies questioned the ecological validity and real-world transferability of VR. Short tasks, narrow measurement windows, and reliance on subjective scales restricted behavioral insight. Sampling biases – driven by repeated-measure designs, technology-enthusiastic participants, and predominantly student populations – further limited generalizability. Constrained stimulus sets, limited modalities, and occasional multilingual data inconsistencies added complexity.  \textbf{Sample size and representativeness} (\textit{n=14}) posed additional challenges: many studies relied on small or pilot samples, with demographic imbalances (e.g., male-dominated, culturally homogeneous groups) and low XR experience producing novelty effects and limiting causal inference. Challenges also involved limited \textbf{interaction depth and multimodality} (\textit{n=10}), i.e., robotic systems that usually only have one modality of interaction and therefore appear to be less adaptive and dynamic. Many studies relied on a single interaction channel despite participant demand for richer multimodal cues. Brief encounters prevented social bonding and reduced realism, while proxemics and gaze inconsistencies diminished authenticity. Technical constraints occasionally biased interaction logs, limiting ecological validity. Finally, studies often featured \textbf{short durations and a lack of longitudinal data} (\textit{n=8}). 


\textbf{Future work} aligns closely with these limitations, emphasizing \textbf{technical advancements} (\textit{n=19}). Researchers call for better sensing and tracking (e.g., improved eye/hand tracking, biosignals), more stable networks, and lighter head-mounted displays. Additional priorities include optimized graphical user interfaces, adaptive gesture generation, natural voice interaction, and support for complex cooperative tasks. Multi-user XR highlights the importance of synchronous interactions between multiple humans and robots. \textbf{Deployment studies} (\textit{n=15}) are recommended to align simulation and real-world outcomes. Testing in authentic environments (homes, schools, clinics) is considered essential for uncovering ecological factors such as lighting, noise, and connectivity that remain hidden in lab-based VR. \textbf{Longitudinal studies} (\textit{n=12}) are proposed to examine long-term learning, habituation, trust formation, and preference stability. Such designs are crucial for determining whether gains in motivation, competence, and trust endure over time and for deriving reliable design parameters for future robot systems. \textbf{Social and emotional interaction} (\textit{n=11}) is another key direction. Researchers seek to enhance robot nonverbal behaviors – gestures, emotion expression, gaze – while studying dialogue quality, trust-building, and relationship dynamics. Moving beyond task performance, the goal is to model authentic social bonding in XR. Improving \textbf{measurements and metrics} (\textit{n=8}), including validated immersion scales, multidimensional pupil analyses, and longitudinal protocols, is considered essential for integrating technical, social, and methodological perspectives. Greater emphasis on \textbf{diverse populations and cultural factors} (\textit{n=7}) was mentioned in order to overcome the pronounced WEIRD bias and to understand culturally shaped interaction norms. Finally, \textbf{adaptive and personalized systems} (\textit{n=5}) and \textbf{multimodal interaction} (\textit{n=5}) are highlighted as mutually reinforcing directions. Real-time adaptation to user cognition, emotion, and context – potentially via reinforcement learning – depends on integrating multiple communication channels (speech, gestures, gaze). Together, these measures support richer and more accurate user modeling.

\begin{table}[htbp]
\centering
\scriptsize
\setlength{\tabcolsep}{2.5pt}
\renewcommand{\arraystretch}{0.93}

\caption{Overview of codes and results for RQ3: \emph{Challenges} and \emph{Future Work}.}
\label{tab:challenges_futurework_xr_hri}

\begin{tabularx}{\linewidth}{@{} l l c >{\raggedright\arraybackslash}X @{}}
\toprule
\textbf{Code} & \textbf{Subcategory} & \textbf{Count} & \textbf{Paper}\\
\midrule

\multicolumn{4}{@{}l}{\textbf{Challenges}}\\
\addlinespace[1pt]

& \textbf{Technical Limitations} & \textbf{21} & \\
& \hspace{1em}Network / synchronization & 4 &
\cite{Fujii2020CoEatingMR,Gao2019MetaRLTrustHRI,Nigro2024InteractiveARProxemics,Strassmann2024DontJudgeBook}\\
& \hspace{1em}Sim-to-real gap & 3 &
\cite{Nertinger2024RobotPostureSafety,Petrak2019ProxemicAwarenessFirstImpressions,Pilacinski2023RobotEyesTrustPerformance}\\
& \hspace{1em}Wizard-of-Oz overhead & 3 &
\cite{Helgert2025aVRStudyTool,Schulten2025MergingRealities,Zhiyu2024ArchitecturalRoboticsRestorative}\\
& \hspace{1em}Device constraints & 3 &
\cite{Brown2023MixedRealityDeictic,Fujii2020CoEatingMR,Groechel2019ExpressiveMRAms}\\
& \hspace{1em}Tracking / sensing accuracy & 3 &
\cite{Gao2019MetaRLTrustHRI,Groechel2023MoveToCodeAR,Zhang2025PredictingPerceptionsNavigation}\\
& \hspace{1em}Lack of haptic feedback & 2 &
\cite{Mueller2017RoboticWorkmate,Peters2018SocialDistancesMR}\\
& \hspace{1em}UI / interface issues & 2 &
\cite{Groechel2023MoveToCodeAR,Helgert2025aVRStudyTool}\\
& \hspace{1em}System stability \& crashes & 1 &
\cite{Groechel2021KinestheticCuriosity}\\
\addlinespace[3pt]

& \textbf{Methodological Limitations} & \textbf{21} & \\
& \hspace{1em}Ecological validity & 8 &
\cite{Helgert2024UnderstandableTransparency,Mueller2017RoboticWorkmate,Nertinger2024RobotPostureSafety,Peters2018SocialDistancesMR,Petrak2019ProxemicAwarenessFirstImpressions,Sadka2020VRSimulationNonHumanoid,Wijnen2020HRIUserStudiesVR,Zhiyu2024ArchitecturalRoboticsRestorative}\\
& \hspace{1em}Short tasks & 3 &
\cite{Fujii2020CoEatingMR,Gao2019MetaRLTrustHRI,Groechel2019ExpressiveMRAms}\\
& \hspace{1em}Sampling / participant bias & 3 &
\cite{Karakosta2023SAROptimizingAR,Kim2019CollaborativeVRGameTeens,Wang2019EmbodimentSubstrateDecisionMaking}\\
& \hspace{1em}Stimulus / interaction constraints & 2 &
\cite{Groechel2019ExpressiveMRAms,Nigro2024InteractiveARProxemics}\\
& \hspace{1em}Language / dataset mismatch & 2 &
\cite{Mizuchi2022VRGUIBehaviorCollection,Mueller2017RoboticWorkmate}\\
& \hspace{1em}Limited autonomy & 2 &
\cite{Karakosta2023SAROptimizingAR,Wijnen2020HRIUserStudiesVR}\\
& \hspace{1em}Novelty \& immersion effects & 1 &
\cite{Zhiyu2024ArchitecturalRoboticsRestorative}\\
\addlinespace[3pt]

& \textbf{Sample Size \& Representativeness} & \textbf{14} & \\
& \hspace{1em}Small / pilot samples & 7 &
\cite{Brown2023MixedRealityDeictic,Groechel2021KinestheticCuriosity,Karakosta2023SAROptimizingAR,Mahajan2020UsabilityMetricsMR,Nigro2024InteractiveARProxemics,Schulten2025MergingRealities,Zhiyu2024ArchitecturalRoboticsRestorative}\\
& \hspace{1em}Demographic imbalance & 2 &
\cite{Bjorling2022IAmTheRobot,Nigro2024InteractiveARProxemics}\\
& \hspace{1em}Cultural homogeneity & 2 &
\cite{Chenlin2023PerspectiveTakingProsocial,Fujii2020CoEatingMR}\\
& \hspace{1em}Lack of control groups & 2 &
\cite{Karakosta2023SAROptimizingAR,Schulten2025MergingRealities}\\
& \hspace{1em}Low XR experience & 1 &
\cite{Gao2019MetaRLTrustHRI}\\
\addlinespace[3pt]

& \textbf{Interaction Depth \& Multimodality} & \textbf{10} & \\
& \hspace{1em}Reduced interaction depth & 4 &
\cite{Kim2019CollaborativeVRGameTeens,Pozharliev2021AttachmentStyleResponses,Schulten2025MergingRealities,Ye2022RCareWorld}\\
& \hspace{1em}Limited modalities & 3 &
\cite{Helgert2024UnderstandableTransparency,Mizuchi2022VRGUIBehaviorCollection,Wang2019EmbodimentSubstrateDecisionMaking}\\
& \hspace{1em}Proxemics \& realism issues & 2 &
\cite{Peters2018SocialDistancesMR,Pilacinski2023RobotEyesTrustPerformance}\\
& \hspace{1em}Bias \& data quality & 1 &
\cite{Mizuchi2022VRGUIBehaviorCollection}\\
\addlinespace[3pt]

& \textbf{Short Duration \& Longitudinality} & \textbf{8} & \\
& \hspace{1em}Short overall duration & 3 &
\cite{Arpaia2022WearableBCI,Chenlin2023PerspectiveTakingProsocial,Fan2022SARConnect}\\
& \hspace{1em}No long-term follow-ups & 2 &
\cite{Mueller2017RoboticWorkmate,Pilacinski2023RobotEyesTrustPerformance}\\
& \hspace{1em}Mini-tasks / micro experiments & 2 &
\cite{Fujii2020CoEatingMR,Gao2019MetaRLTrustHRI}\\
& \hspace{1em}Repetition effects in short timeframe & 1 &
\cite{Karakosta2023SAROptimizingAR}\\

\midrule
\multicolumn{4}{@{}l}{\textbf{Future Work}}\\
\addlinespace[1pt]

& \textbf{Technical Enhancements} & \textbf{19} & \\
& \hspace{1em}Sensors \& tracking & 7 &
\cite{Arpaia2022WearableBCI,Brown2023MixedRealityDeictic,Groechel2021KinestheticCuriosity,Lee2017AppearanceTherapyEngagement,Nertinger2024RobotPostureSafety,Strassmann2024DontJudgeBook,Zhang2025PredictingPerceptionsNavigation}\\
& \hspace{1em}Display \& hardware & 4 &
\cite{Fujii2020CoEatingMR,Li2018PostureEmbodimentSocialDistance,Schulten2025MergingRealities,Wang2019EmbodimentSubstrateDecisionMaking}\\
& \hspace{1em}Interface \& usability & 4 &
\cite{Groechel2023MoveToCodeAR,Helgert2025aVRStudyTool,Mizuchi2022VRGUIBehaviorCollection,Nigro2024InteractiveARProxemics}\\
& \hspace{1em}Communication \& dialog & 2 &
\cite{Wijnen2020HRIUserStudiesVR,Pilacinski2023RobotEyesTrustPerformance}\\
& \hspace{1em}Collaboration \& multi-user XR & 2 &
\cite{Kim2019CollaborativeVRGameTeens,Ye2022RCareWorld}\\
\addlinespace[3pt]

& \textbf{Deployment Studies} & \textbf{15} & \\
& \hspace{1em}Sim-to-real validation & 8 &
\cite{Nertinger2024RobotPostureSafety,Peters2018SocialDistancesMR,Petrak2019ProxemicAwarenessFirstImpressions,Pilacinski2023RobotEyesTrustPerformance,Pozharliev2021AttachmentStyleResponses,Schulten2025MergingRealities,Wijnen2020HRIUserStudiesVR,Zhiyu2024ArchitecturalRoboticsRestorative}\\
& \hspace{1em}Real-world deployments \& context & 7 &
\cite{Brown2023MixedRealityDeictic,Mizuchi2022VRGUIBehaviorCollection,Fan2022SARConnect,Groechel2021KinestheticCuriosity,Groechel2019ExpressiveMRAms,Lee2017AppearanceTherapyEngagement,Sadka2020VRSimulationNonHumanoid}\\
\addlinespace[3pt]

& \textbf{Longitudinal Studies} & \textbf{12} & \\
& \hspace{1em}Long-term interventions & 8 &
\cite{Arpaia2022WearableBCI,Bjorling2022IAmTheRobot,Mahajan2020UsabilityMetricsMR,Nertinger2024RobotPostureSafety,Schulten2025MergingRealities,Chenlin2023PerspectiveTakingProsocial,Karakosta2023SAROptimizingAR,Pozharliev2021AttachmentStyleResponses}\\
& \hspace{1em}Acceptance \& preference stability & 4 &
\cite{Mizuchi2022VRGUIBehaviorCollection,Nigro2024InteractiveARProxemics,Petrak2019ProxemicAwarenessFirstImpressions,Pilacinski2023RobotEyesTrustPerformance}\\
\addlinespace[3pt]

& \textbf{Social \& Emotional Interaction} & \textbf{11} & \\
& \hspace{1em}Gestures, emotion \& nonverbal behavior & 6 &
\cite{Bjorling2022IAmTheRobot,Li2018PostureEmbodimentSocialDistance,Lee2017AppearanceTherapyEngagement,Peters2018SocialDistancesMR,Pilacinski2023RobotEyesTrustPerformance,Zhiyu2024ArchitecturalRoboticsRestorative}\\
& \hspace{1em}Dialogue, trust \& relationship & 5 &
\cite{Brown2023MixedRealityDeictic,Fujii2020CoEatingMR,Gao2019MetaRLTrustHRI,Nigro2024InteractiveARProxemics,Sadka2020VRSimulationNonHumanoid}\\
\addlinespace[3pt]

& \textbf{Improved Measurement \& Metrics} & \textbf{8} &
\cite{Chenlin2023PerspectiveTakingProsocial,Gao2019MetaRLTrustHRI,Helgert2025aVRStudyTool,Lee2017AppearanceTherapyEngagement,Mahajan2020UsabilityMetricsMR,Pilacinski2023RobotEyesTrustPerformance,Pozharliev2021AttachmentStyleResponses,Zhiyu2024ArchitecturalRoboticsRestorative}\\
\addlinespace[3pt]

& \textbf{Cultural Factors} & \textbf{7} &
\cite{Chenlin2023PerspectiveTakingProsocial,Fujii2020CoEatingMR,Helgert2024UnderstandableTransparency,Lee2017AppearanceTherapyEngagement,Li2018PostureEmbodimentSocialDistance,Schulten2025MergingRealities,Strassmann2024DontJudgeBook}\\
\addlinespace[3pt]

& \textbf{Adaptive \& Personalized Systems} & \textbf{5} &
\cite{Fan2022SARConnect,Groechel2023MoveToCodeAR,Karakosta2023SAROptimizingAR,Mueller2017RoboticWorkmate,Petrak2019ProxemicAwarenessFirstImpressions}\\
\addlinespace[3pt]

& \textbf{Multimodal Interaction} & \textbf{5} &
\cite{Gao2019MetaRLTrustHRI,Helgert2024UnderstandableTransparency,Kim2019CollaborativeVRGameTeens,Nigro2024InteractiveARProxemics,Nertinger2024RobotPostureSafety}\\

\bottomrule
\end{tabularx}
\end{table}

\subsection{RQ4: Characteristics of Researchers and Participants}

A total of 162 \textbf{researchers} were identified as authors. The binary sex distribution is uneven, with 92 men and 70 women. Since sex was inferred rather than self-reported, it refers to observable sex rather than gender identity. In terms of the background of researchers involved, social XR-HRI is dominated by technology-oriented disciplines, which account for roughly 70\% of author backgrounds (\textit{n=81}). Interdisciplinary, psychological, and other non-technical fields (e.g., education, medicine, design, law) appear only sporadically (\textit{n=43}). The field is strongly shaped by early-career scholars: PhD candidates (\textit{n} = 58) and students (\textit{n} = 34) make up the majority, supported by substantial senior involvement from professors (\textit{n} = 57). Only three contributors are software engineers.  Geographically, the field is concentrated in Western countries – led by the USA (\textit{n} = 13) and Germany (\textit{n} = 8) – with smaller contributions from several European countries. Asian participation is scattered, and the Global South is nearly absent, leaving culture-specific interaction norms underrepresented.

A similar pattern emerges on the side of \textbf{participants} (overall age: \textit{M} = 31.16, \textit{SD} = 20.82). The sample is predominantly male (490 vs. 335 female, 3 diverse), with 255 unspecified cases. Most participants are university students (\textit{n} = 616), making recruitment convenient but unrepresentative for real-world user groups. Children (\textit{n} = 53), teenagers (\textit{n} = 38), and older adults (\textit{n} = 42) appear only when tied to application domains such as education or healthcare. Age distribution ranges widely – from primary school age to over 75 – but median values fall within the student bracket. Study locations mirror this bias: most sessions took place in university laboratories (\textit{n} = 653), compared to far fewer in public spaces (\textit{n} = 76) or school settings (\textit{n} = 73). Documentation gaps remain substantial: nearly one-third of participant information reports (\textit{n} = 266) lacks details on identity, sex/originx educational status, origin, or age.

\section{Discussion}

The present literature review shows that XR technologies are currently used primarily as controllable laboratory instruments in HRI research, with a focus on evaluation and simulation. While this focus enables precise experimental control and cost-effective iteration, it comes at the expense of ecological validity and long-term practical transfer. Domains such as public space, industry, and medical facilities are currently severely underrepresented, even though these contexts are likely to be crucial for the integration of social robots. Technical and methodological bottlenecks characterize the status quo. On a technical level, limitations in head-mounted displays, tracking accuracy, network stability, and a lack of haptic feedback hinder realism. However, these obstacles are expected to be resolved with the advent of new XR devices. Methodologically, short lab sessions, small homogeneous samples, and self-reported measurements dominate. Objective sensor technology and automated behavioral analysis remain the exception. These limitations indicate that social XR-HRI results are often only partially generalizable. Researchers should critically reflect on their reasons for using XR in research while considering the specific benefits the medium provides. Another key shortcoming is the lack of diversity – both in the research teams and among the participants. Recent work has concluded that limited sample diversity can lead to blind spots in HRI research \cite{Seaborn2023DiversityHRI,erlefocusgroupalgbias2025}. On the researcher team side, increased diversity can lead to improved performance \cite{Horwitz} and reduce bias against marginalized groups \cite{PettigrewTropp2006ContactMetaAnalysis}. The majority of studies are conducted in Western-influenced, technology-oriented contexts with predominantly young adults. Children, older adults, and people from non-Western cultures are rarely found in the samples. Interdisciplinary approaches that systematically integrate social, psychological, or ethical perspectives are likely rare, presumably due to their limited number. In terms of content, most studies operate at an interaction level that is stimulus-driven rather than dialogical. Language is frequently used, but almost exclusively in script-based form. Multimodal, adaptively responsive robot interactions remain a research shortcoming. Particularly striking is the lack of important reporting, e.g., on the hardware and software used, but also on robots. These are essential for working in accordance with scientific ethics, but also for conducting reproducible research. A driving factor is that, to date, there is no uniform XR taxonomy for describing such specifications. The use of XR is described in many ways, but rarely classified with precision. Existing taxonomy approaches (e.g.,~\cite{ADRIANACARDENASROBLEDO2022101863}) offer valuable insights, but have not yet been tailored to the specifics of HRI. The need for a domain-specific, social XR-HRI taxonomy that clearly structures roles, contexts, and forms of use is particularly noteworthy, as it would offer researchers a clear classification option/framework.

\section{Future Roadmap for Social XR-HRI}

We propose a four-part roadmap for XR as a research instrument for social HRI in order to establish the medium as a reliable research and transfer instrument for social robots:
\begin{enumerate}

    \item \textbf{Strengthen Application Contexts}. Many social XR-HRI studies are still conducted in digital laboratory settings rather than in specific application contexts. This means that robots and their interactions are often viewed in isolation. Yet, social HRI is shaped by context and may not translate well from simplified lab reproductions \cite{10.3389/fpsyg.2015.00637}. Researchers should take advantage of the greater ecological validity of XR and conduct their studies in realistic application environments, using their own creations or ready-made scenes (e.g., from the Unity Store). This approach also promotes a certain degree of replicability, as virtual environments can be shared with the community.

    \item \textbf{Build More Robust and Testable Technological Iterations}. The present review shows that current social XR-HRI systems often remain proof-of-concept prototypes, with WoZ control and rudimentary interactions of the robot (e.g., scripted dialogue). While this reflects the very high implementation effort required for XR applications and the pressure of short publication cycles in scientific research, the field would benefit from longer-term, iterative systems that go beyond “flat” demos. Sustainable foundations must be laid, e.g., through toolkits, modular XR–robot integration frameworks, and authoring tools with reusable interaction modules. This process could reduce overhead and improve robustness and reproducibility.

    \item \textbf{Embed Diversity in Samples and Research Teams}. Social XR-HRI studies often rely on WEIRD, technology-oriented participant samples and similarly homogeneous research teams, which limits generalizability and carries the risk of systematic bias \cite{Seaborn2023DiversityHRI}. These circumstances naturally pose a challenge for the entire research landscape, not just social XR-HRI. Nevertheless, researchers should expand recruitment across cultures, age groups, genders, and technical experience levels, reflecting diversity in team composition and participatory design practices. We argue that a key lever is stronger collaboration across labs, disciplines, and sectors – especially through international collaborations that can provide local contextual expertise, improve cultural validity, and enable access to participant groups that are difficult to reach. Furthermore, awareness of diversity issues is a cornerstone that needs to be strengthened among researchers in order to increase their willingness to include a broad spectrum of people in their own studies.

    \item \textbf{Build Reporting Standards}. A key problem in XR-HRI research is the lack of reporting standards that capture both technical details and HRI-relevant aspects. While conceptual taxonomies exist, they mainly categorize systems and interaction paradigms rather than defining what must be reported for transparency and reproducibility \cite{Walker}. The SLR shows that crucial information is often incomplete or omitted. This lack of transparency undermines reproducibility and interpretability, making it difficult to judge whether effects stem from interaction design, robot behavior, measurement choices, or implementation differences. Because XR setups and robot platforms are highly heterogeneous, a multidimensional reporting standard is needed, covering XR roles, technical factors, contextual factors, and methodological characteristics. A practical solution could be a checklist-based reporting framework (PRISMA-style) tailored to XR-HRI. Standardized reporting would improve cross-study comparability, enable replication, and make study designs, robot behaviors, and data pipelines more transparent and reusable.

\end{enumerate}

\section{Limitations}

The SLR has several limitations. First, it should be noted that the search strategy only covered four databases, even if these are major publication venues for HRI and have a certain technical focus. Second, the coding process involves subjective interpretations, especially in the few inductive categories. However, qualitative coding aims to derive concepts and categories through interpretation rather than purely objective classification. Third, the review focuses on studies published since 2015; while this corresponds to the availability of modern XR systems, it ignores earlier and most recent work. Fourth, this review is limited in its precision of details because the examined studies lack information, particularly on the demographic characteristics of the participants, the hardware/software specifications, and the methodological details in some categories. Furthermore, the order of authors listed in publications does not reveal who contributed how much to the individual works. This could distort the evaluation in terms of the distribution of disciplines. Fifth, literature reviews are often subject to publication bias, meaning that most studies with significant results are published, while others may be omitted. This could be exacerbated in the field of XR because the medium is relatively new, highly experimental, and complex. The review presented may therefore not cover all studies that would normally have been relevant.

\section{Conclusion}

This literature review provides an overview of how XR has been used as a research tool in social robotics over the past ten years. The results show that the unique possibilities of XR are not being fully exploited. Rigid and unimodal robot interactions often dominate, with robots frequently presented as objects to be observed in laboratory studies, limited use of AR and MR, low participant diversity, and a strong reliance on self-reported measurements. While these approaches offer a high degree of experimental control, they come at the expense of ecological validity, inclusivity, and transferability to the real world. In particular, technical and methodological bottlenecks have been identified, as well as inconsistent reporting practices that prevent reproducibility and comparability. To take social XR-HRI a step further towards being a recognized research tool, future work should focus on long-term studies, place more trust in the functional capabilities of virtual robots, promote diversity among researchers – but above all among participants – and create a standardized social XR-HRI taxonomy. Doing so will not only enhance the scientific robustness of social XR-HRI research but also ensure that its insights and applications are relevant, transferable, and socially inclusive.

\begin{credits}
\subsubsection{\ackname} The presented work was supported by the RuhrBots competence center (16SV8693) funded by the German Federal Ministry of Research, Technology and Space. We would like to thank Alexandra-Michelle Drechsler, Lukas Erle and Ramón-Darius Imort for providing feedback on the manuscript. 

\subsubsection{\discintname}
The authors have no competing interests to declare that are relevant to the content of this article.
\end{credits}
%
%
%
%


\end{document}